\documentclass[10pt,twocolumn,a4paper]{article}

\usepackage{spconf}
\usepackage{times}
\usepackage{epsfig}
\usepackage{graphicx}
\usepackage{amsmath}
\usepackage{amssymb}
\usepackage{booktabs}

\usepackage{enumitem}

\usepackage{xcolor}

\usepackage{siunitx}
\DeclareMathOperator*{\argmin}{arg\,min} 
\newcommand{\cash}[1]{\SI{#1}[\$]{}}

\begin{document}

\title{Fine-Grained Property Value Assessment using Probabilistic Disaggregation}

                \name{
                Cohen Archbold$^1$
                \hspace{1cm}
                Benjamin Brodie$^2$
                \hspace{1cm}
                Aram Ansary Ogholbake$^1$
                \hspace{1cm}
                Nathan Jacobs$^3$}
                \address{$^1$University of Kentucky
                \hspace{1cm}
                $^2$BlueHalo
                \hspace{1cm}
                $^3$Washington University in St. Louis}

\maketitle
\thispagestyle{empty}

\begin{abstract}

The monetary value of a given piece of real estate, a parcel, is often readily available from a geographic information system. However, for many applications, such as insurance and urban planning, it is useful to have estimates of property value at much higher spatial resolutions.~We propose a method to estimate the distribution over property value at the pixel level from remote sensing imagery.~We evaluate on a real-world dataset of a major urban area. Our results show that the proposed approaches are capable of generating 
fine-level estimates of property values, significantly improving upon a diverse collection of baseline approaches.


\end{abstract}

\section{Introduction}

Many remote-sensing applications often consist of estimating the value of a quantity at a high spatial resolution. This includes quantities such as population density estimation or species distribution modeling~\cite{jacobs2018weakly}. While not true for every application, it is often the case that labeled training data is only available at a lower spatial resolution than the resolution at which we would like to obtain estimates. This makes training a model to make high spatial resolution predictions challenging. We propose an approach to overcoming this challenge in a way that is useful across a broad range of applications.

We address the task of per-pixel regression of values where the labels available are aggregated over spatial regions. Previous work on this topic~\cite{jacobs2018weakly} used a region aggregation layer to estimate population density at the pixel level based on satellite imagery and census data. A key limitation of this method is that it provides a direct point estimate without information about uncertainty or confidence of predictions. We quantify pixel and region values with probability distributions. 

We provide the theoretical motivation and demonstrate the effectiveness of value disaggregation using an analytic approach, where fine-level distributions are grouped to create a coarse-level distribution.



The main application of the probabilistic disaggregation method that we consider is identification of the visual features from overhead imagery that contribute and detract the most from the monetary value of a parcel. To facilitate this, we have collected a dataset consisting of publicly accessible aerial imagery and property values from Hennepin County, Minnesota, which includes the urban area of Minneapolis and the surrounding mix of industrial, suburban, and rural areas. We demonstrate the ability to accurately predict parcel values based on a pixel-level prediction of contributions to value. 

The key contributions of this work include: (1) providing the theoretical motivation for a probabilistic disaggregation method for fine-level estimation from coarse labels; (2) formulating a method for probabilistic disaggregation and demonstrating its quantitative and qualitative value; and (3) Showing how the method works on property value estimation from overhead imagery.




\section{Related Work}

\textbf{Weakly Supervised Segmentation \& Object Counting:} Our work produces pixel-level value distribution predictions from labels that are aggregated over large image regions and is closely related to works in image segmentation and object counting. Recently, many powerful CNN-based object detectors such as YOLO~\cite{yolo} and R-CNN~\cite{Girshick2014RichFH} present high detection accuracy on objects in sparse scenes. 

Regression-based counting has proved a more effective method for crowd density estimation in noisier and denser crowds~\cite{gao2020cnnbased}. Gao et al.~\cite{gao2020cnnbased} estimates pixel-level density and a count is obtained by integrating over the region.  Our method follows this last approach, in which we estimate the pixel-level contributions to property value and sum them to create estimates over wider regions.
In remote sensing, there is a scarcity of data for object-counting tasks. One approach is generating pseudo pixel-level labels~\cite{Gao2020remotesensing}.
For this work, we focus on using weak labels for regression or counting tasks, rather than segmentation, in overhead imagery. 
We extend previous work~\cite{jacobs2018weakly} which uses coarse-grained labels to predict density functions at pixel level. 

\textbf{Probabilistic Representation:}
  Probabilistic representation of internal features and/or model outputs has proven effective in a variety of machine learning tasks. For example, Variational Autoencoders (VAE)~\cite{Kingma2014AutoEncodingVB}, embed features in a probabilistic latent space. Samples from the distribution are then passed through a decoder to reconstruct the original input. A key development that allows probabilistic features to be used in machine learning applications is the reparameterization trick~\cite{Kingma2014AutoEncodingVB}, which provides a differentiable way of sampling from a variety of probability distributions.

\section{Problem Statement}

We address the task of estimating pixel-level property values using fine-level aerial imagery and parcel-level property values. Property value labels exist at the parcel-level, but we seek to quantify the contribution to the overall value of each pixel. Since property values are often used in downstream processing tasks, such as insurance risk adjustment, it is critical to also provide a quantification of the uncertainty in these estimates.

We model this as the problem of estimating a geospatial function, $v(l)\in\mathbb{R}^+$, for a location, $l$ on the surface of the earth, from overhead imagery, $I$, of the location. This function, $v$, reflects the value of the property, including the land and building value, at a particular location. We redefine this as a gridded geospatial function, $v(p_i)$, where the value corresponds to the value in the area imaged by a pixel, $p_i$. Therefore, the problem reduces to making pixel-level predictions of $v$ from the input imagery.

The key challenge is that we do not have samples from the function, $v$.  We only have a set of spatially aggregated values, $Y=\left\{ y_1\ldots y_n\right\}$, where each value, $y_i \in \mathbb{R}^+$, represents the value of the corresponding parcel/regions, $R=\left\{ r_1\ldots r_n\right\}$. Specifically, we define $y_i = V(r_i) = \sum _{p_j\in r_i}v\left( p_j\right)$. These regions could have arbitrary topology and be overlapping, but in practice will typically be simply connected and disjoint.

As discussed above, the labels are provided at the parcel/region level. To generate the estimated values for each parcel, we assume the value $y_i$ of the region $r_i$ is distributed according to:
\begin{equation}
\label{eq:prob_formulation}
    y_i \sim \sum _{p_j \in r_i}w_i(p_j) P(v(p_j)|I(p_j); \Theta) 
\end{equation}
where $w_i(p_j)$ is value at pixel $p_j$ that belongs to region $r_i$ and $P(v(p_j)|I(p_j);  \Theta)$ is a distribution for the value of pixel $p_j$ given image of location $p_j$ and the parameters $\Theta$.

Previous work has shown that it is possible to train a model, in a weakly supervised fashion, to make pixel-level predictions using only these aggregated labels~\cite{jacobs2018weakly}. The key observation was that since the aggregation process is linear, it is possible to propagate derivatives through the operation, enabling end-to-end optimization of a neural network that predicts $v$. 

The previous work does not capture uncertainty in the underlying predictions, which limits its practical usefulness. We estimate pixel-level and parcel-level Gaussian probability distributions for the value of the underlying location.

\section{Approach}

We propose to use a CNN, which takes the overhead imagery as input, to generate a pixel-level probabilistic value map. We model pixel-level values as Gaussian distributions with the mean representing the most likely value of for the corresponding piece of land and variance modeling uncertainty in the estimation. For simplicity, we model the output pixel-value distributions as independent and allow the CNN to capture dependencies between pixel distributions. By aggregating pixel-level distributions for each pixel in a parcel, we obtain an estimate for the overall value of the parcel, for which we have labels.



The sum of independent Gaussian distributions can be written analytically as follows: given $X_1 \sim \mathcal{N}(\mu_1,\,\sigma^{2}_1)\,$ and $X_2 \sim \mathcal{N}(\mu_2,\,\sigma^{2}_2)\,$, then $X_1+X_2 \sim \mathcal{N}(\mu_1+\mu_2,\,\sigma^{2}_1+\sigma^{2}_2)\,$. 

Therefore, given a mean $\mu$ and and variance $\sigma^{2}$ for the value prediction at each pixel, the aggregated observation on a region $r_i$ is: 
\begin{equation}\label{eq:aggregated_gaussian}
    \hat{y_i}(\Theta) \sim \mathsf{\mathcal{N}}\left(\sum_{p_j\in r_i}\hat{\mu}( p_j; \Theta),\sum_{p_j\in r_i}\hat{\sigma^{2}}( p_j; \Theta)\right).
\end{equation}

We optimize the system by minimizing the negative log-likelihood of the the parcel value label according the aggregated estimate distribution $\hat{y_i}(\Theta)$. That is, given $\mu_i^\star = \sum_{p_j\in r_i}\hat{\mu}( p_j; \Theta)$ and ${\sigma_i^\star}^2 = \sum_{p_j\in r_i}\hat{\sigma^{2}}( p_j; \Theta)$ for each region $r_i$, we find the parameters $\Theta$ that minimize the negative log-likelihood of the resulting Gaussian:
\begin{equation}\label{eq:nll}
    \argmin_\Theta \sum_{i} -\log\left( 
     \frac{1}{\sqrt{2\pi}\sigma_i^\star} \exp\left(-\frac{1}{2} \frac{\left(y_i - \mu_i^\star\right)^2}{{\sigma_i^\star}^2} \right)\right),
\end{equation}
given the ground truth value $y_i$.

\subsection{Baselines}
\label{sec:baselines}
We compare the results of the probabilistic disaggregation method against two deep learning baselines: a uniform value assumption and a linear aggregation method.

With the uniform value assumption, we assume that the value of each parcel is evenly distributed among the pixels included in the parcel. This results in strong pixel-level labels for all viable pixels in the dataset, although these strong labels may not be the best representation of the data. Using these labels, we train a probabilistic backbone to predict pixel-level distributions matching the pixel-level labels.

\begin{figure*}
        \centering
        \includegraphics[width=0.88\linewidth]{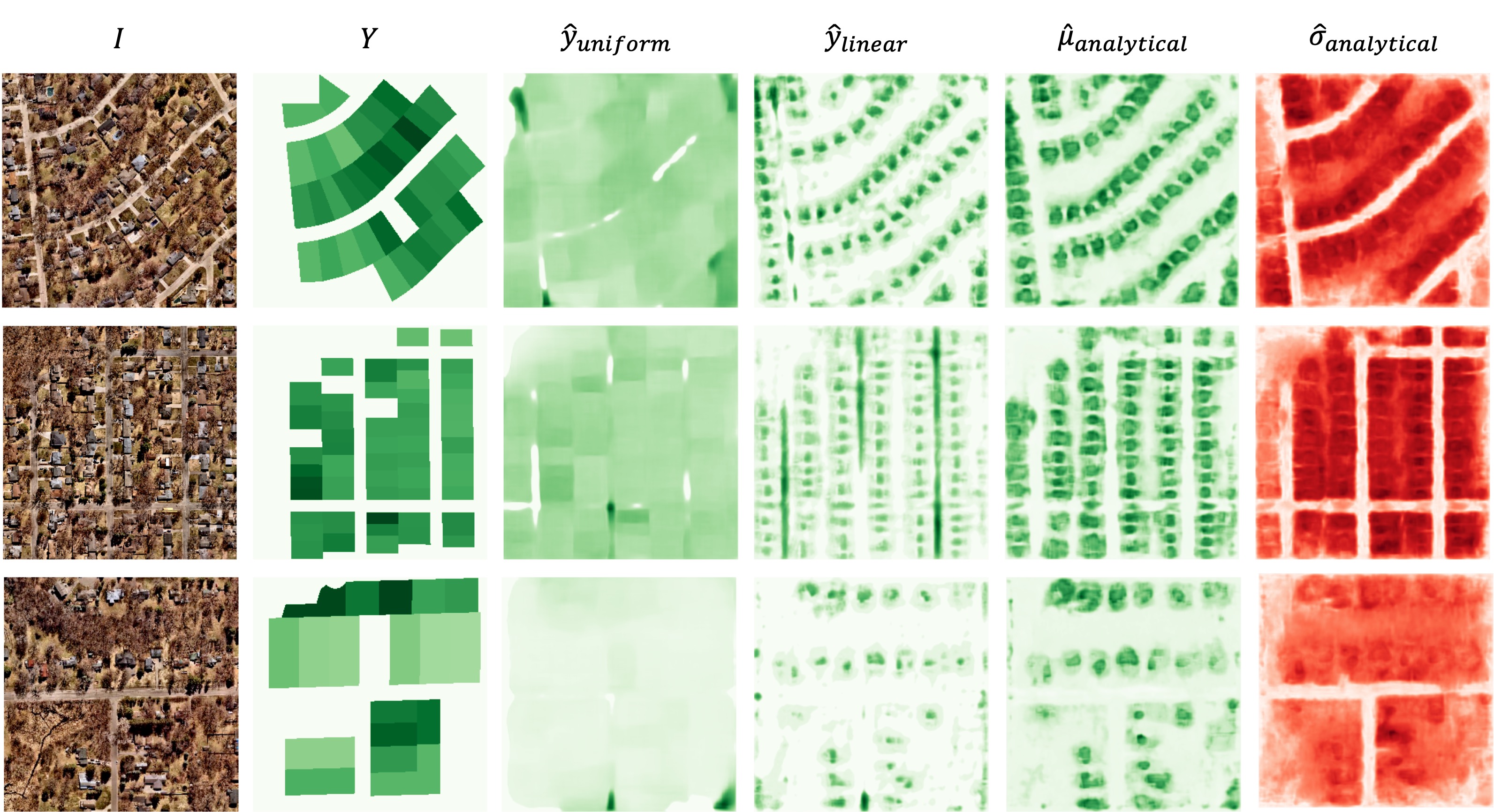}
        \caption{Qualitative examples of mean value predictions of the baselines described in Section~\ref{sec:baselines} and disaggregation model performing on the Hennepin County Dataset~\ref{sec:hennepin}. We show the original input imagery $I$ and parcel regions $Y$ shaded by corresponding value(Columns 1-2). We observe that building pixels are valued more highly than lawn pixels using the disaggregation methods (Columns 3-5). As well, we show the standard deviation predictions of the dissaggregation model (Column 6).}
        \label{fig:qualitative_results}
    \end{figure*}

We also use a deterministic linear aggregation as a baseline for comparison to the probabilistic method. This method treats the input as a per pixel value-estimation, and sums the pixel values over the regions using a single linear layer. We then optimize the baseline with Mean Squared Error (MSE) for the pixel value estimate. This method is essentially the method presented in~\cite{jacobs2018weakly}, which was applied to the similar problem of population density estimation.

As well, in Table \ref{tab:merged_results} we compare the methods against a Gaussian fit of the training set labels. We expect performance in this case to be trivially attainable.

\section{Evaluation}
\label{sec:evaluation}
 We estimate the contribution to overall property value of each pixel(square meter) that is visible from overhead imagery in the Hennepin County Dataset. At test-time, we take the means $\mu$ of the probabilistic models as the per-pixel value estimation. We aggregate the $\hat{y}$ of the respective methods and compare to the true value.

\begin{table*}
\centering
{
\setlength{\tabcolsep}{1em}
\renewcommand{\arraystretch}{1.2}
 \begin{tabular}{lllllll} 
 \toprule
 Method  & MAE & MAPE & $\mathcal{P}_\pm(10^4)$ & $\mathcal{P}_\pm(10^5)$ & Average $\hat{\sigma}$ & Log Prob (avg) \\ [0.5ex]
 \midrule
 Gaussian Fit & \cash{271280} & $126.15\% $  & $1.48\%$ & $14.74\%$ & \cash{436840} & $-14.12$ \\ 
 
 Uniform Value & \cash{177574}  & $81.59\%$ & $11.11\%$  & $79.44\%$  & \cash{2588} & $-2062.19$ \\
  
 Deterministic & \cash{61093}  & $25.66\%$ & NA & NA & NA & NA\\ 
 
 \midrule
 
 
 
 
 Analytical & \cash{63311} & $27.14\%$ & $14.5\%$ & $84.38\%$ & \cash{3547} & $-317.70$ \\
 
 \bottomrule
\end{tabular}
}
\caption{Results of disaggregation on the Hennepin County Dataset with random merging of neighboring parcels during training. We show Mean Absolute Error (MAE) and Mean Absolute Percentage Error (MAPE) of predictions with ground truth parcel values. Results of the analytical approach to probabilistic disaggregation are compared to the two baselines described in Section~\ref{sec:baselines} as well as Gaussian Fit. We also present the probability of being within \cash{10000} and \cash{100000} of the ground truth value according to the predicted Gaussian distribution, along with the average predicted standard deviation.}
\label{tab:merged_results}
\end{table*}
 
\subsection{Hennepin County Dataset}
\label{sec:hennepin}

We built an evaluation dataset using the imagery and parcel information made available by Hennepin County, Minnesota. We randomly selected 1915 sub-regions and extracted $302 \times 302$ chips with a ground-sample distance of \SI{1}{\meter} from the 2020 aerial imagery. The corresponding parcel geometries and values (accurate as of 2020) were extracted from GIS Open Data. For each chip, only parcels fully contained within the chip were retained. In addition, we removed outliers and other defects from the collected dataset. For instance, many of the original parcels included a market value of zero, often for either unlabeled or public land. As well, we remove parcels with disproportionately high values from downtown Minneapolis. 

Each sample of the dataset contains the information: region image, parcel masks, and corresponding market values. We hold out 10\% of the dataset for validation and 10\% of the dataset for testing.


\subsection{Implementation Details}

We implemented the proposed approach in PyTorch and trained each model on an NVIDIA A6000 GPU. We use UNet~\cite{ronneberger2015u} as a backbone, which provides an output map at the same resolution as the input image. The network produces a Gaussian probability distribution for each input pixel by generating a two channel output with the first channel corresponding to the means and the second channel corresponding to the variances. We apply the softplus function, $f(x) = \log(1+ \exp{(x)})$, to the output of the second channel to ensure positivity of the predicted variances, $\hat{\sigma}^2$. To compute the aggregation of parcel pixels, we model the computation after \cite{jacobs2018weakly} using a region aggregation layer. This computation necessitates memory sufficient to represent a matrix $M \in \mathbb{R}^{n \times HW}$, where $n$ is the number of regions and $HW$ is the flattened image size. Due to this, we are limited to a batch size of 8. We train each model for \num{300} epochs with a learning rate of \num{1e-3} under the Adam optimization algorithm. We select the best checkpoint based on performance on the validation set, and present results on the test set

\subsection{Spatial Region Combination Experiment}


A major challenge of the experiments on parcel value is the lack of pixel-level ground truth labels. We propose an experiment to evaluate the ability of a model trained on coarse labels to predict on fine labels. To accomplish this, we modify the training and validation set to combine neighboring parcels for each region. Corresponding parcel values for joined regions are summed and we skip any region combinations that exceed \SI{1000}[\$] per $m^2$. At inference, we evaluate on single parcels as in the standard experiment. 
We indicate that the disaggregation method is able to accurately estimate pixel-level values by demonstrating that we effectively estimate single parcel values from merged parcel labels, as shown in Table~\ref{tab:merged_results}. That is, we are able to predict on finer-level labels than the set of labels that we trained on. The probabilistic method performs on par with the deterministic baselines, while again providing uncertainty estimations.




\section{Conclusion}
 
We presented a method for estimating pixel-level distributions of property values from remote sensing imagery. This is a challenging task to evaluate because pixel-level ground truth is infeasible to collect. We addressed this by training on datasets with artificially merged parcels and evaluating on the unmerged parcels. While this doesn't directly evaluate the pixel-level distributions, it does show that our method is able to determine the spatial distribution of property values more accurately than baseline approaches. These results demonstrate that our approach significantly improves upon several strong baseline approaches. While not demonstrated in this work, we believe probabilistic disaggregation can be applied to numerous applications where fine-level, probabilistic estimates are needed but fine-level training data is difficult to obtain, such as population density estimation and species distribution modeling.

\section*{Acknowledgements}

This research is based upon work supported in part by the Office of the Director of National Intelligence (ODNI), Intelligence Advanced Research Projects Activity (IARPA), via Contract Number 2021-20111000005. The views and conclusions contained herein are those of the authors and should not be interpreted as necessarily representing the official policies, either expressed or implied, of ODNI, IARPA, or the U.S. Government. The U.S. Government is authorized to reproduce and distribute reprints for governmental purposes notwithstanding any copyright annotation therein.

{\small
\bibliographystyle{ieee_fullname}
\bibliography{biblio}
}

\end{document}